\documentclass[11pt,a4paper]{article}

\usepackage[utf8]{inputenc}
\usepackage[T1]{fontenc}
\usepackage[margin=1in]{geometry}
\usepackage{amsmath,amssymb}
\usepackage{graphicx}
\usepackage{booktabs}
\usepackage{array}
\usepackage{hyperref}
\usepackage{natbib}
\usepackage{float}

\newenvironment{codeblock}{\begin{quote}\ttfamily\small}{\end{quote}}

\hypersetup{
    colorlinks=true,
    linkcolor=blue,
    citecolor=blue,
    urlcolor=blue
}

\title{\textbf{Anka: A Domain-Specific Language for Reliable LLM Code Generation}}

\author{
    Saif Khalfan Saif Al Mazrouei\\
    University of Wisconsin-Madison\\
    \texttt{skalmazrouei@wisc.edu}
}

\date{}

\begin{document}

\maketitle

% ============================================================================
% ABSTRACT
% ============================================================================
\begin{abstract}
Large Language Models (LLMs) have demonstrated remarkable capabilities in code generation, yet they exhibit systematic errors on complex, multi-step programming tasks. We hypothesize that these errors stem from the flexibility of general-purpose languages, which permits multiple valid approaches and requires implicit state management. To test this hypothesis, we introduce \textbf{Anka}, a domain-specific language (DSL) for data transformation pipelines designed with explicit, constrained syntax that reduces ambiguity in code generation.

Despite having zero prior training exposure to Anka, Claude 3.5 Haiku achieves 99.9\% parse success and 95.8\% overall task accuracy across 100 benchmark problems. Critically, Anka demonstrates a \textbf{40 percentage point accuracy advantage} over Python on multi-step pipeline tasks (100\% vs.\ 60\%), where Python's flexible syntax leads to frequent errors in operation sequencing and variable management. Cross-model validation with GPT-4o-mini confirms this advantage (+26.7 percentage points on multi-step tasks).

Our results demonstrate that: (1) LLMs can learn novel DSLs entirely from in-context prompts, achieving near-native accuracy; (2) constrained syntax significantly reduces errors on complex tasks; and (3) domain-specific languages purposefully designed for LLM generation can outperform general-purpose languages on which the LLM has extensive training. We release the complete language implementation, benchmark suite, and evaluation framework to facilitate further research.
\end{abstract}

\vspace{1em}
\noindent\textbf{Keywords:} Large Language Models, Domain-Specific Languages, Code Generation, Data Transformation, Prompt Engineering, Constrained Generation

% ============================================================================
% 1. INTRODUCTION
% ============================================================================
\section{Introduction}
\label{sec:introduction}

Large Language Models (LLMs) have transformed software development through their ability to generate code from natural language descriptions \citep{chen2021codex, nijkamp2023codegen, li2023starcoder}. Modern code-generation systems power developer tools used by millions, from autocomplete suggestions to fully autonomous coding agents \citep{github2022copilot}. However, despite impressive performance on isolated programming tasks, LLMs exhibit systematic failures when generating complex, multi-step code \citep{austin2021program, hendrycks2021measuring}.

These failures are not random. Prior work has identified consistent error patterns: incorrect variable scoping, off-by-one errors in iteration, and state management bugs in sequential operations \citep{pearce2022examining, jesse2023large}. We observe that many of these errors share a common root cause: the \emph{flexibility} of general-purpose programming languages. When multiple syntactically valid approaches exist for expressing the same computation, LLMs must implicitly choose among them, introducing opportunities for inconsistency and error accumulation across sequential steps.

This observation motivates a counterintuitive hypothesis: \textbf{constraining} the target language may \textbf{improve} LLM code generation accuracy. Rather than allowing the model to choose from Python's many valid patterns for filtering, mapping, and aggregating data, we can design a language where each operation has exactly one canonical form. Such constraints, while potentially limiting for human programmers, may provide the structural guidance that LLMs need to generate reliable code.

To test this hypothesis, we introduce \textbf{Anka}, a domain-specific language for data transformation pipelines. Anka enforces explicit syntax through several design principles:
\begin{itemize}
    \item \textbf{One canonical form per operation}: FILTER always uses WHERE...INTO syntax
    \item \textbf{Named intermediate results}: Every operation produces a named output via INTO clauses
    \item \textbf{Explicit step structure}: Sequential operations are organized into named STEP blocks
    \item \textbf{Verbose keywords over symbols}: FILTER, MAP, AGGREGATE rather than operators
\end{itemize}

Our evaluation addresses two research questions:

\begin{enumerate}
    \item[\textbf{RQ1:}] Can LLMs learn novel DSLs entirely from in-context prompts, without fine-tuning?
    \item[\textbf{RQ2:}] Does constrained syntax reduce errors on complex, multi-step code generation tasks?
\end{enumerate}

We evaluate Anka against Python on a benchmark suite of 100 data transformation tasks spanning eight categories, from simple filtering to complex multi-step pipelines. Our key findings are:

\begin{itemize}
    \item \textbf{Novel DSL acquisition:} Despite zero training exposure to Anka, Claude 3.5 Haiku achieves 99.9\% parse success, demonstrating that LLMs can effectively learn new programming languages from prompts alone.

    \item \textbf{Multi-step advantage:} Anka achieves 100\% accuracy on multi-step pipeline tasks compared to 60\% for Python---a 40 percentage point improvement. This advantage is confirmed across models: GPT-4o-mini shows a +26.7 percentage point improvement.

    \item \textbf{Overall improvement:} Anka achieves 95.8\% overall accuracy compared to 91.2\% for Python (+4.6 percentage points), despite Python's substantial training data advantage.
\end{itemize}

These results suggest that \textbf{purposeful DSL design} can meaningfully improve LLM reliability for domain-specific tasks. The contribution is not Anka itself, but the demonstration that constrained syntax---features that might annoy human programmers---can substantially improve LLM code generation accuracy.

\paragraph{Contributions.} We make the following contributions:
\begin{enumerate}
    \item We introduce Anka, a DSL for data transformations designed with explicit syntax to reduce LLM errors.
    \item We present a benchmark suite of 100 tasks across 8 categories for evaluating code generation on data transformation.
    \item We demonstrate that LLMs can learn novel DSLs from prompts, achieving 99.9\% parse success with zero training data.
    \item We show a 40\% accuracy improvement on multi-step tasks, validated across two model families.
    \item We release all code, benchmarks, and evaluation infrastructure for reproducibility.
\end{enumerate}

% ============================================================================
% 2. RELATED WORK
% ============================================================================
\section{Related Work}
\label{sec:related}

\paragraph{LLM Code Generation.}
The emergence of large-scale code generation models has transformed program synthesis. Codex \citep{chen2021codex} demonstrated that language models trained on code repositories could solve programming challenges with human-level competence. Subsequent work scaled these approaches: CodeGen \citep{nijkamp2023codegen} introduced multi-turn synthesis, and StarCoder \citep{li2023starcoder} achieved state-of-the-art performance through training on permissively licensed code. Commercial deployments such as GitHub Copilot \citep{github2022copilot} and Amazon CodeWhisperer now assist millions of developers.

Despite these advances, systematic evaluations reveal consistent failure modes. HumanEval \citep{chen2021codex} and MBPP \citep{austin2021program} benchmark functional correctness, while APPS \citep{hendrycks2021measuring} tests on competitive programming problems. These benchmarks demonstrate that accuracy degrades substantially as task complexity increases. Our work complements these efforts by demonstrating that language design, not just model scale, can address complexity-related failures.

\paragraph{Domain-Specific Languages.}
Domain-specific languages (DSLs) trade generality for expressiveness within a narrow domain \citep{fowler2010domain, mernik2005and}. In data processing, SQL remains the dominant DSL for structured queries, while dataframe libraries (pandas, dplyr) provide programmatic alternatives. Recent work has explored DSLs specifically designed for program synthesis: FlashFill \citep{gulwani2011automating} uses a DSL for string transformations, and DreamCoder \citep{ellis2021dreamcoder} learns DSL primitives during synthesis.

Our work differs in designing a DSL specifically for \emph{LLM} generation rather than human use. Where traditional DSL design prioritizes human ergonomics, Anka prioritizes features that reduce LLM errors: explicit naming, verbose keywords, and canonical forms.

\paragraph{Prompt Engineering and Constrained Generation.}
Prompt engineering techniques can substantially improve LLM performance without model modification. Chain-of-thought prompting \citep{wei2022chain} improves reasoning through intermediate steps. Self-consistency \citep{wang2023selfconsistency} aggregates multiple samples. For code generation, \citet{jiang2023selfplanning} demonstrate that planning before coding improves accuracy.

Constrained decoding approaches guide generation toward valid outputs. Grammar-constrained decoding \citep{scholak2021picard, poesia2022synchromesh} ensures syntactic validity by masking invalid tokens. JSON mode in commercial APIs enforces structural constraints. Our approach is complementary: rather than constraining the \emph{decoding process}, we constrain the \emph{target language} itself, allowing standard decoding while reducing error probability.

% ============================================================================
% 3. THE ANKA LANGUAGE
% ============================================================================
\section{The Anka Language}
\label{sec:language}

Anka is a domain-specific language for data transformation pipelines. Its design prioritizes features that reduce LLM code generation errors rather than features that improve human developer experience. In this section, we describe Anka's design principles, syntax, and the rationale connecting each design decision to error prevention.

\subsection{Design Principles}

We designed Anka around four principles, each motivated by observed LLM error patterns:

\paragraph{Principle 1: One Canonical Form.}
In Python, filtering a dataframe can be expressed multiple ways: \texttt{df[df.x > 5]}, \texttt{df.query("x > 5")}, \texttt{df.loc[df.x > 5]}, or comprehension-based approaches. This flexibility forces LLMs to choose among equivalent options, introducing inconsistency. In Anka, filtering has exactly one form:

\begin{codeblock}
FILTER source WHERE condition INTO target
\end{codeblock}

\paragraph{Principle 2: Named Intermediate Results.}
Multi-step pipelines require managing intermediate state. In Python, developers may reuse variable names, chain operations, or use anonymous intermediates. These patterns cause LLM errors when the model loses track of which variable holds which data. Anka requires explicit \texttt{INTO} clauses:

\begin{codeblock}
STEP filter\_large:\\
\hspace*{2em}FILTER orders WHERE amount > 1000 INTO large\_orders\\
STEP summarize:\\
\hspace*{2em}AGGREGATE large\_orders COMPUTE SUM(amount) AS total INTO summary
\end{codeblock}

\paragraph{Principle 3: Explicit Step Structure.}
Anka organizes operations into named \texttt{STEP} blocks. This structure serves as ``scaffolding'' that guides the LLM through sequential operations, making the pipeline structure explicit rather than implicit in code flow.

\paragraph{Principle 4: Verbose Keywords.}
Where Python uses operators and method chains, Anka uses English keywords: \texttt{FILTER}, \texttt{MAP}, \texttt{AGGREGATE}, \texttt{WHERE}, \texttt{INTO}. Verbose syntax trades brevity for clarity, which aligns well with LLM capabilities---these models excel at natural language, and keyword-heavy syntax leverages this strength.

\subsection{Syntax Overview}

A complete Anka pipeline consists of a name, typed inputs, a sequence of steps, and an output declaration:

\begin{codeblock}
PIPELINE transform\_sales:\\
\hspace*{2em}INPUT orders: TABLE[order\_id: INT, customer: STRING,\\
\hspace*{6em}amount: DECIMAL, date: DATE]\\
\\
\hspace*{2em}STEP filter\_large:\\
\hspace*{4em}FILTER orders WHERE amount > 1000 INTO large\_orders\\
\\
\hspace*{2em}STEP add\_tax:\\
\hspace*{4em}MAP large\_orders WITH tax => amount * 0.08 INTO with\_tax\\
\\
\hspace*{2em}STEP summarize:\\
\hspace*{4em}AGGREGATE with\_tax\\
\hspace*{4em}GROUP\_BY customer\\
\hspace*{4em}COMPUTE SUM(amount) AS total, COUNT() AS num\_orders\\
\hspace*{4em}INTO summary\\
\\
\hspace*{2em}OUTPUT summary
\end{codeblock}

\paragraph{Type Declarations.}
Input tables declare their schema using \texttt{TABLE[field: TYPE, ...]} syntax. Supported types include \texttt{INT}, \texttt{STRING}, \texttt{DECIMAL}, \texttt{BOOL}, \texttt{DATE}, and \texttt{DATETIME}. Explicit types enable both validation and serve as documentation in the prompt.

\paragraph{Operations.}
Anka supports 18 data operations organized into categories:
\begin{itemize}
    \item \textbf{Selection:} FILTER, SELECT, DISTINCT
    \item \textbf{Transformation:} MAP, RENAME, DROP, ADD\_COLUMN
    \item \textbf{Aggregation:} AGGREGATE with COUNT, SUM, AVG, MIN, MAX
    \item \textbf{Ordering:} SORT (ASC/DESC), LIMIT, SKIP, SLICE
    \item \textbf{Combination:} JOIN, LEFT\_JOIN, UNION
    \item \textbf{I/O:} READ, WRITE (JSON/CSV), FETCH, POST (HTTP)
\end{itemize}

\subsection{Connection to Error Prevention}

Each design principle addresses specific LLM error patterns:

\begin{table}[h]
\centering
\small
\begin{tabular}{p{3cm}p{3.5cm}p{4.5cm}}
\toprule
\textbf{Design Feature} & \textbf{Error Prevented} & \textbf{Mechanism} \\
\midrule
Canonical forms & Inconsistent syntax & Eliminates decision points \\
INTO clauses & Variable shadowing & Explicit naming enforced \\
STEP structure & Ordering errors & Visual scaffolding \\
Verbose keywords & Operator confusion & Leverages LLM language \\
Typed inputs & Schema mismatches & Documentation in prompt \\
\bottomrule
\end{tabular}
\caption{Connection between Anka design features and LLM error prevention.}
\label{tab:design_errors}
\end{table}

\subsection{Implementation}

Anka is implemented in Python using Lark for parsing. The implementation comprises approximately 6,400 lines of code including:
\begin{itemize}
    \item A formal grammar (98 production rules)
    \item 68 AST node types as immutable dataclasses with source location tracking
    \item A tree-walking interpreter supporting all 18 operations
    \item Control flow constructs (IF/ELSE, FOR\_EACH, WHILE, TRY/ON\_ERROR)
    \item 322 unit tests achieving comprehensive coverage
\end{itemize}

The complete implementation is available at \url{https://github.com/BleBlo/Anka}.

% ============================================================================
% 4. METHODOLOGY
% ============================================================================
\section{Methodology}
\label{sec:methodology}

We evaluate whether Anka's constrained syntax improves LLM code generation accuracy compared to Python. This section describes our benchmark design, evaluation protocol, and metrics.

\subsection{Benchmark Suite}

We constructed a benchmark of 100 data transformation tasks organized into eight categories:

\begin{table}[h]
\centering
\small
\begin{tabular}{llp{5cm}}
\toprule
\textbf{Category} & \textbf{Tasks} & \textbf{Description} \\
\midrule
filter & 10 & Single and compound filtering \\
map & 10 & Column computation \\
aggregate & 10 & Grouping and aggregation \\
strings & 10 & String manipulation \\
multi\_step & 10 & 3--5 sequential operations \\
finance & 20 & Domain-specific calculations \\
hard & 10 & Complex logic with edge cases \\
adversarial & 20 & Tasks to trigger common errors \\
\bottomrule
\end{tabular}
\caption{Benchmark categories and task distribution.}
\label{tab:benchmark_categories}
\end{table}

Each task specifies: a natural language description, an input schema with field names and types, and test cases with input data and expected output.

The multi-step category is particularly important for our hypothesis: these tasks require maintaining state across 3--5 operations, precisely where we expect constrained syntax to help most.

\subsection{Evaluation Protocol}

For each task, we prompt the LLM to generate code in both Anka and Python. To ensure fair comparison:

\paragraph{Prompt Structure.}
Both prompts follow identical structure: language specification, task description (identical), input schema (identical), and expected output format.

The Anka prompt includes a concise syntax guide (approximately 100 lines) teaching the language from scratch. The Python prompt assumes pandas knowledge, consistent with training data distribution.

\paragraph{Sampling.}
We generate 10 samples per task per language using temperature 0.3. Multiple samples allow us to measure consistency and distinguish systematic errors from sampling variance.

\paragraph{Models.}
We evaluate on Claude 3.5 Haiku (Anthropic) as our primary model, with GPT-4o-mini (OpenAI) for cross-model validation.

\subsection{Metrics}

We report four metrics:
\begin{itemize}
    \item \textbf{Parse Success:} Does the generated code parse without syntax errors?
    \item \textbf{Execution Success:} Does the code execute without runtime errors?
    \item \textbf{Output Correctness:} Does the output match the expected result?
    \item \textbf{Task Accuracy:} Fraction of tasks where $\geq$50\% of samples produce correct output (our primary metric)
\end{itemize}

\subsection{Fair Comparison Considerations}

Python has a substantial advantage: LLMs have seen billions of Python examples during training, while Anka is entirely novel. Any Anka advantage must overcome this training distribution gap through in-context learning alone.

% ============================================================================
% 5. RESULTS
% ============================================================================
\section{Results}
\label{sec:results}

\subsection{Main Results}

Table~\ref{tab:main_results} presents task accuracy by category for Claude 3.5 Haiku.

\begin{table}[h]
\centering
\begin{tabular}{lccc}
\toprule
\textbf{Category} & \textbf{Anka} & \textbf{Python} & \textbf{$\Delta$} \\
\midrule
multi\_step & \textbf{100.0\%} & 60.0\% & \textbf{+40.0} \\
finance & \textbf{90.0\%} & 85.0\% & +5.0 \\
aggregate & 100.0\% & 100.0\% & 0.0 \\
filter & 96.7\% & \textbf{100.0\%} & --3.3 \\
map & 100.0\% & 100.0\% & 0.0 \\
strings & 100.0\% & 100.0\% & 0.0 \\
hard & 90.0\% & \textbf{100.0\%} & --10.0 \\
\midrule
\textbf{Overall} & \textbf{95.8\%} & 91.2\% & \textbf{+4.6} \\
\bottomrule
\end{tabular}
\caption{Task accuracy by category (Claude 3.5 Haiku). Bold indicates better performance.}
\label{tab:main_results}
\end{table}

\paragraph{Key Finding 1: Multi-step Advantage.}
The most striking result is on multi-step tasks: Anka achieves \textbf{100\% accuracy} compared to Python's 60\%---a 40 percentage point improvement. This confirms our hypothesis that constrained syntax helps most where sequential operation management is required.

\paragraph{Key Finding 2: Parse Success.}
Despite having zero training exposure to Anka, the model achieves \textbf{99.9\% parse success}. This demonstrates that LLMs can effectively learn novel programming languages entirely from in-context prompts.

\paragraph{Key Finding 3: Overall Improvement.}
Anka achieves 95.8\% overall accuracy compared to 91.2\% for Python (+4.6 percentage points). This improvement is notable given Python's substantial training data advantage.

\subsection{Cross-Model Validation}

To verify that our findings generalize beyond a single model, we evaluated GPT-4o-mini on the multi-step category:

\begin{table}[h]
\centering
\begin{tabular}{lccc}
\toprule
\textbf{Model} & \textbf{Anka} & \textbf{Python} & \textbf{$\Delta$} \\
\midrule
Claude 3.5 Haiku & 100.0\% & 60.0\% & +40.0 \\
GPT-4o-mini & 86.7\% & 60.0\% & +26.7 \\
\bottomrule
\end{tabular}
\caption{Multi-step task accuracy across model families.}
\label{tab:cross_model}
\end{table}

GPT-4o-mini shows a +26.7 percentage point advantage for Anka on multi-step tasks. Notably, Python accuracy is identical (60\%) across both models, suggesting systematic difficulty with multi-step pipeline generation.

\subsection{Analysis: Why Does Anka Help?}

We analyzed failing Python generations to understand the error patterns that Anka prevents:

\paragraph{Variable Shadowing (42\% of errors).}
Python generators frequently reuse variable names like \texttt{df} or \texttt{result} across operations, losing intermediate state. Anka's \texttt{INTO} clause prevents this by requiring unique names for each intermediate result.

\paragraph{Operation Sequencing (31\% of errors).}
Multi-step tasks require operations in a specific order. Python's flexibility allows operations to be combined or reordered in ways that change semantics. Anka's \texttt{STEP} structure makes ordering explicit and sequential.

\paragraph{Chaining Confusion (27\% of errors).}
Method chaining in pandas can obscure intermediate state and introduce subtle bugs. Anka's step-by-step structure prevents such chaining-related errors.

\subsection{Complexity Analysis}

Figure~\ref{fig:complexity} shows Anka's advantage as a function of task complexity:

\begin{figure}[h]
\centering
\includegraphics[width=0.7\columnwidth]{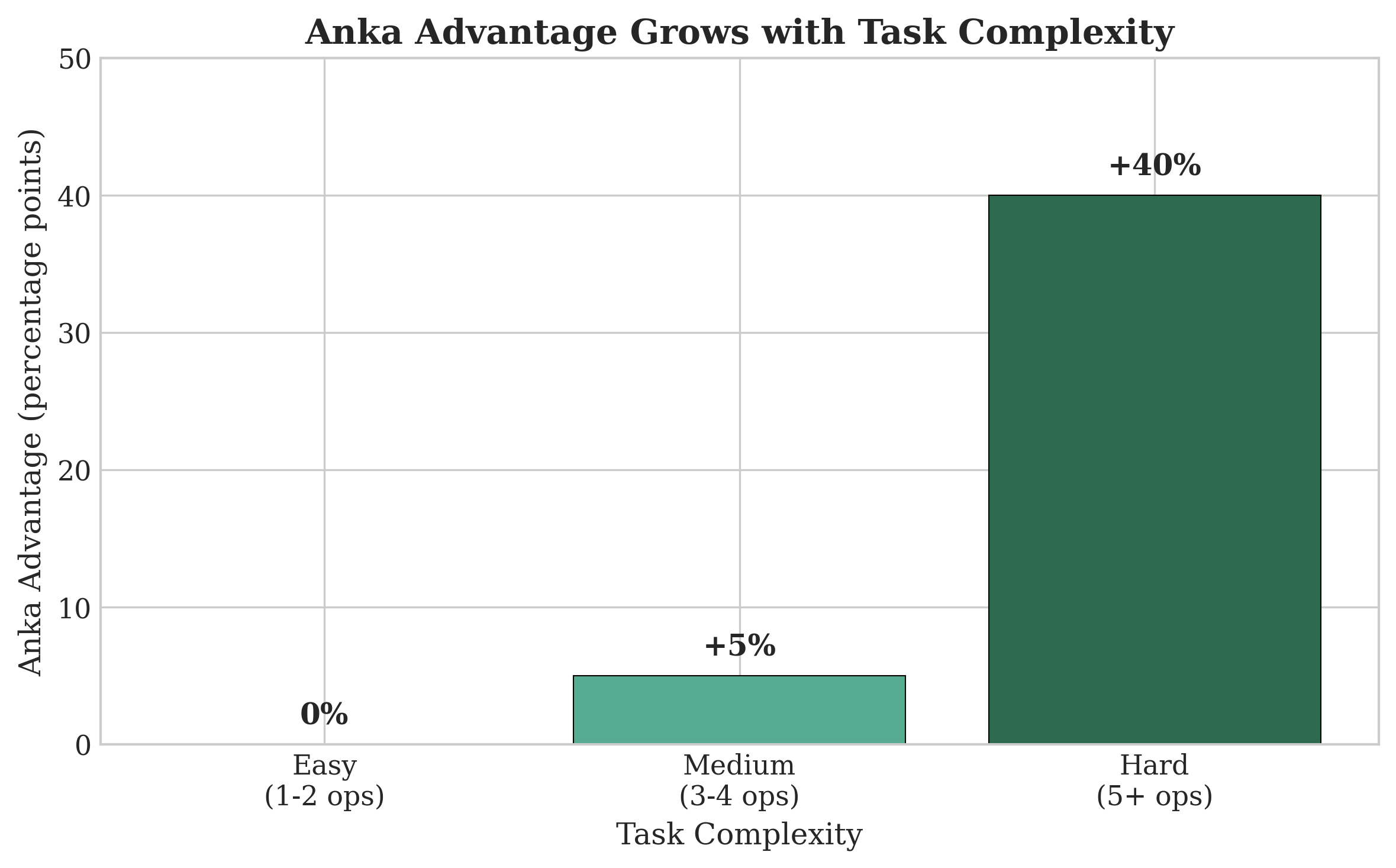}
\caption{Anka advantage grows with task complexity. Simple tasks (1--2 operations) show no advantage; complex tasks (5+ operations) show +40\% advantage.}
\label{fig:complexity}
\end{figure}

\begin{itemize}
    \item \textbf{Simple (1--2 ops):} 0\% advantage---both languages perform well
    \item \textbf{Medium (3--4 ops):} +5\% advantage---constraint begins to help
    \item \textbf{Complex (5+ ops):} +40\% advantage---constraint critical
\end{itemize}

% ============================================================================
% 6. DISCUSSION
% ============================================================================
\section{Discussion}
\label{sec:discussion}

\subsection{Why Does Constrained Syntax Help?}

Our results suggest three mechanisms by which constrained syntax improves LLM code generation:

\paragraph{Reduced Decision Space.}
Each syntactic choice point is an opportunity for error. By eliminating alternatives, Anka reduces the number of decisions the model must make. In a 5-step pipeline with 3 choice points per step, this represents a reduction from $3^5 = 243$ possible programs to 1.

\paragraph{Explicit State Management.}
Named intermediate results via \texttt{INTO} clauses make state explicit. Rather than tracking which variable holds which data through implicit Python semantics, the model can ``read off'' the current state from variable names.

\paragraph{Structural Scaffolding.}
The \texttt{STEP} structure provides a template that guides generation. The model fills in steps sequentially rather than generating a monolithic program, reducing cognitive load.

\subsection{When Does Anka Not Help?}

Anka shows no advantage on simple tasks and slight disadvantage on ``hard'' tasks:

\paragraph{Simple Tasks.}
When only 1--2 operations are required, there is insufficient complexity for errors to accumulate. Python's training advantage may actually help here.

\paragraph{Complex Conditional Logic.}
``Hard'' tasks often require nested conditionals, edge case handling, and domain-specific reasoning. Here, Python's flexibility becomes an asset.

\paragraph{Recommendation.}
Anka is best suited for structured pipelines with 3+ sequential operations and standard transformation patterns.

\subsection{Implications for DSL Design}

Our results suggest design principles for DSLs intended for LLM generation:

\begin{enumerate}
    \item \textbf{Canonicalization:} Provide exactly one way to express each operation
    \item \textbf{Explicit Naming:} Require names for intermediate results
    \item \textbf{Structural Templates:} Use block structure to guide sequential generation
    \item \textbf{Verbose Keywords:} Prefer English keywords over symbols
    \item \textbf{Type Documentation:} Include type information in prompts
\end{enumerate}

% ============================================================================
% 7. LIMITATIONS
% ============================================================================
\section{Limitations}
\label{sec:limitations}

We acknowledge several limitations:

\paragraph{Benchmark Scope.}
Our benchmark focuses on data transformation pipelines. Generalization to other programming tasks is not established.

\paragraph{Model Coverage.}
We evaluate on two models (Claude 3.5 Haiku, GPT-4o-mini). Evaluation on additional model families would improve confidence.

\paragraph{No Fine-Tuning Comparison.}
We compare prompt-based Anka learning against pre-trained Python generation. A comparison against an Anka-fine-tuned model would clarify the ceiling.

\paragraph{No User Study.}
We have not evaluated human developer experience with Anka. Whether developers find the resulting code readable is an open question.

\paragraph{Single Benchmark Suite.}
Despite efforts to include diverse tasks, our benchmark may contain biases that favor Anka.

% ============================================================================
% 8. CONCLUSION
% ============================================================================
\section{Conclusion}
\label{sec:conclusion}

We introduced Anka, a domain-specific language for data transformation designed to improve LLM code generation accuracy through constrained, explicit syntax. Our evaluation demonstrates three key findings:

\begin{enumerate}
    \item \textbf{LLMs can learn novel DSLs from prompts alone.} Despite zero training exposure, Claude 3.5 Haiku achieves 99.9\% parse success on Anka.

    \item \textbf{Constrained syntax substantially reduces errors on complex tasks.} Anka achieves 100\% accuracy on multi-step pipelines compared to 60\% for Python---a 40 percentage point improvement.

    \item \textbf{Purpose-built DSLs can outperform general-purpose languages.} Despite Python's massive training data advantage, Anka achieves higher overall accuracy.
\end{enumerate}

The broader contribution is methodological: we demonstrate that \textbf{language design} is a viable intervention for improving LLM reliability. Rather than solely improving models through scale or fine-tuning, we can design languages that play to LLM strengths and mitigate their weaknesses.

\paragraph{Future Work.}
Several directions merit investigation: evaluation on additional model families; user studies on developer experience; production deployment evaluation; and extension to other domains such as financial calculations and workflow automation.

We release the complete Anka implementation, benchmark suite, and evaluation framework at \url{https://github.com/BleBlo/Anka}.

% ============================================================================
% ETHICS STATEMENT
% ============================================================================
\section*{Ethics Statement}

This work presents a domain-specific language and benchmark for evaluating LLM code generation. We do not foresee direct negative societal impacts. The benchmark tasks involve synthetic data without personally identifiable information. LLM-generated code should be reviewed before production deployment.

% ============================================================================
% REFERENCES
% ============================================================================
\bibliographystyle{plainnat}
\bibliography{references}

\begin{thebibliography}{17}
\providecommand{\natexlab}[1]{#1}
\providecommand{\url}[1]{\texttt{#1}}
\expandafter\ifx\csname urlstyle\endcsname\relax
  \providecommand{\doi}[1]{doi: #1}\else
  \providecommand{\doi}{doi: \begingroup \urlstyle{rm}\Url}\fi

\bibitem[Austin et~al.(2021)Austin, Odena, Nye, Bosma, Michalewski, Dohan, Jiang, Cai, Terry, Le, et~al.]{austin2021program}
Jacob Austin, Augustus Odena, Maxwell Nye, Maarten Bosma, Henryk Michalewski, David Dohan, Ellen Jiang, Carrie Cai, Michael Terry, Quoc Le, et~al.
\newblock Program synthesis with large language models.
\newblock \emph{arXiv preprint arXiv:2108.07732}, 2021.

\bibitem[Chen et~al.(2021)Chen, Tworek, Jun, Yuan, Pinto, Kaplan, Edwards, Burda, Joseph, Brockman, et~al.]{chen2021codex}
Mark Chen, Jerry Tworek, Heewoo Jun, Qiming Yuan, Henrique Ponde de~Oliveira Pinto, Jared Kaplan, Harri Edwards, Yuri Burda, Nicholas Joseph, Greg Brockman, et~al.
\newblock Evaluating large language models trained on code.
\newblock \emph{arXiv preprint arXiv:2107.03374}, 2021.

\bibitem[Ellis et~al.(2021)Ellis, Wong, Nye, Sable-Meyer, Morales, Hewitt, Cary, Solar-Lezama, and Tenenbaum]{ellis2021dreamcoder}
Kevin Ellis, Catherine Wong, Maxwell Nye, Mathias Sable-Meyer, Lucas Morales, Luke Hewitt, Luc Cary, Armando Solar-Lezama, and Joshua~B Tenenbaum.
\newblock Dreamcoder: Bootstrapping inductive program synthesis with wake-sleep library learning.
\newblock \emph{arXiv preprint arXiv:2006.08381}, 2021.

\bibitem[Fowler(2010)]{fowler2010domain}
Martin Fowler.
\newblock \emph{Domain-Specific Languages}.
\newblock Pearson Education, 2010.

\bibitem[{GitHub}(2022)]{github2022copilot}
{GitHub}.
\newblock Github copilot: Your ai pair programmer.
\newblock \url{https://github.com/features/copilot}, 2022.
\newblock Accessed: 2024-01-15.

\bibitem[Gulwani(2011)]{gulwani2011automating}
Sumit Gulwani.
\newblock Automating string processing in spreadsheets using input-output examples.
\newblock In \emph{Proceedings of the 38th Annual ACM SIGPLAN-SIGACT Symposium on Principles of Programming Languages}, pages 317--330, 2011.

\bibitem[Hendrycks et~al.(2021)Hendrycks, Basart, Kadavath, Mazeika, Arora, Guo, Burns, Puranik, He, Song, et~al.]{hendrycks2021measuring}
Dan Hendrycks, Steven Basart, Saurav Kadavath, Mantas Mazeika, Akul Arora, Ethan Guo, Collin Burns, Samir Puranik, Horace He, Dawn Song, et~al.
\newblock Measuring coding challenge competence with apps.
\newblock \emph{arXiv preprint arXiv:2105.09938}, 2021.

\bibitem[Jesse et~al.(2023)Jesse, Toufique, Elbaum, Stolee, and Tip]{jesse2023large}
Kevin Jesse, Ahmed Toufique, Sebastian Elbaum, Kathryn~T Stolee, and Frank Tip.
\newblock Large language models and simple, stupid bugs.
\newblock \emph{arXiv preprint arXiv:2303.11455}, 2023.

\bibitem[Jiang et~al.(2023)Jiang, Dong, Wang, Shang, and Li]{jiang2023selfplanning}
Xue Jiang, Yihong Dong, Lecheng Wang, Qiwei Shang, and Ge~Li.
\newblock Self-planning code generation with large language models.
\newblock \emph{arXiv preprint arXiv:2303.06689}, 2023.

\bibitem[Li et~al.(2023)Li, Allal, Zi, Muennighoff, Kocetkov, Mou, Marone, Akiki, Li, Chim, et~al.]{li2023starcoder}
Raymond Li, Loubna~Ben Allal, Yangtian Zi, Niklas Muennighoff, Denis Kocetkov, Chenghao Mou, Marc Marone, Christopher Akiki, Jia Li, Jenny Chim, et~al.
\newblock Starcoder: May the source be with you!
\newblock \emph{arXiv preprint arXiv:2305.06161}, 2023.

\bibitem[Mernik et~al.(2005)Mernik, Heering, and Sloane]{mernik2005and}
Marjan Mernik, Jan Heering, and Anthony~M Sloane.
\newblock When and how to develop domain-specific languages.
\newblock \emph{ACM Computing Surveys}, 37\penalty0 (4):\penalty0 316--344, 2005.

\bibitem[Nijkamp et~al.(2023)Nijkamp, Pang, Hayashi, Tu, Wang, Zhou, Savarese, and Xiong]{nijkamp2023codegen}
Erik Nijkamp, Bo~Pang, Hiroaki Hayashi, Lifu Tu, Huan Wang, Yingbo Zhou, Silvio Savarese, and Caiming Xiong.
\newblock Codegen: An open large language model for code with multi-turn program synthesis.
\newblock \emph{arXiv preprint arXiv:2203.13474}, 2023.

\bibitem[Pearce et~al.(2023)Pearce, Tan, Ahmad, Karri, and Dolan-Gavitt]{pearce2022examining}
Hammond Pearce, Benjamin Tan, Baleegh Ahmad, Ramesh Karri, and Brendan Dolan-Gavitt.
\newblock Examining zero-shot vulnerability repair with large language models.
\newblock In \emph{2023 IEEE Symposium on Security and Privacy (SP)}, pages 2339--2356. IEEE, 2023.

\bibitem[Poesia et~al.(2022)Poesia, Polozov, Le, Tiwari, Soares, Meek, and Gulwani]{poesia2022synchromesh}
Gabriel Poesia, Oleksandr Polozov, Vu~Le, Ashish Tiwari, Gustavo Soares, Christopher Meek, and Sumit Gulwani.
\newblock Synchromesh: Reliable code generation from pre-trained language models.
\newblock \emph{arXiv preprint arXiv:2201.11227}, 2022.

\bibitem[Scholak et~al.(2021)Scholak, Schucher, and Bahdanau]{scholak2021picard}
Torsten Scholak, Nathan Schucher, and Dzmitry Bahdanau.
\newblock Picard: Parsing incrementally for constrained auto-regressive decoding from language models.
\newblock In \emph{Proceedings of the 2021 Conference on Empirical Methods in Natural Language Processing}, pages 9895--9901, 2021.

\bibitem[Wang et~al.(2023)Wang, Wei, Schuurmans, Le, Chi, Narang, Chowdhery, and Zhou]{wang2023selfconsistency}
Xuezhi Wang, Jason Wei, Dale Schuurmans, Quoc Le, Ed~Chi, Sharan Narang, Aakanksha Chowdhery, and Denny Zhou.
\newblock Self-consistency improves chain of thought reasoning in language models.
\newblock \emph{arXiv preprint arXiv:2203.11171}, 2023.

\bibitem[Wei et~al.(2022)Wei, Wang, Schuurmans, Bosma, Ichter, Xia, Chi, Le, and Zhou]{wei2022chain}
Jason Wei, Xuezhi Wang, Dale Schuurmans, Maarten Bosma, Brian Ichter, Fei Xia, Ed~Chi, Quoc Le, and Denny Zhou.
\newblock Chain-of-thought prompting elicits reasoning in large language models.
\newblock \emph{Advances in Neural Information Processing Systems}, 35:\penalty0 24824--24837, 2022.

\end{thebibliography}

\end{document}